\documentclass[twoside,11pt,preprint]{article}

\usepackage{jmlr2e} 
\usepackage{blindtext}

\usepackage{amsmath} 
\usepackage{tikz} 
\usetikzlibrary{positioning, shapes.geometric, arrows}
\usepackage{listings} 

\usepackage{lastpage}
\jmlrheading{23}{2024}{1-\pageref{LastPage}}{1/24; Revised 5/24}{9/24}{24-0001}{Ruslan Idelfonso Maga\~na Vsevolodovna}

\ShortHeadings{See-Saw Mechanism for Generative AI Code Generation}{Maga\~na Vsevolodovna}
\firstpageno{1}

\begin{document}

\title{See-Saw Generative Mechanism for Scalable Recursive Code Generation with Generative AI}

\author{\name Ruslan Idelfonso Maga\~na Vsevolodovna \email Ruslan.Idelfonso.Magana.Vsevolodovna-cic@ibm.com \\
       \addr IBM Client Innovation Center \\
       Via San Bovio 3, 20054 - Segrate (MI), Italy}

\editor{My Editor}

\maketitle

\begin{abstract}%
The generation of complex, large-scale code projects using generative AI models presents challenges due to token limitations, dependency management, and iterative refinement requirements. This paper introduces the \textit{See-Saw generative mechanism}, a novel methodology for dynamic and recursive code generation. The proposed approach alternates between main code updates and dependency generation to ensure alignment and functionality. By dynamically optimizing token usage and incorporating key elements of the main code into the generation of dependencies, the method enables efficient and scalable code generation for projects requiring hundreds of interdependent files. The mechanism ensures that all code components are synchronized and functional, enabling scalable and efficient project generation. Experimental validation demonstrates the method's capability to manage dependencies effectively while maintaining coherence and minimizing computational overhead.
\end{abstract}

\begin{keywords}
generative AI, recursive code generation, dependency management, scalable systems, token optimization
\end{keywords}

\section{Introduction}
\label{sec:introduction-1}
Generative AI has become a transformative tool in automating software development, offering unprecedented capabilities in code generation. However, its application to large-scale projects remains limited by several critical challenges. One of the foremost issues is the token limitation of large language models (LLMs). These models can only process a limited number of tokens at a time, making it infeasible to generate or validate entire codebases with hundreds of files simultaneously.

Beyond token limitations, the process of generating multiple interdependent files introduces significant complexity. Current generative approaches often treat each file in isolation, lacking awareness of the broader project structure. This locality of generation results in misaligned or incompatible dependencies, leading to inefficient development cycles and the need for extensive manual corrections.

Furthermore, the iterative refinement of generated code exacerbates these inefficiencies. Repeatedly revisiting and updating both main and dependency files to resolve inconsistencies increases computational costs and delays project completion.

These challenges are addressed through the methodology proposed in this paper that not only addresses token and dependency limitations but also ensures coherent and efficient code generation at scale.

Several studies have attempted to tackle the challenges of code generation with varying degrees of success. The work of \cite{xu2022codegen} introduced a transformer-based model capable of generating code across multiple programming languages. While their approach demonstrated impressive results in handling multi-language codebases, it did not address the issue of aligning interdependent files dynamically, leaving room for inconsistencies in large projects.

The study by \cite{brown2020language} showcased the capabilities of GPT-3 for code generation, highlighting its potential to generate syntactically correct code snippets. However, their work also revealed the limitations of such models in handling larger contexts, as the token limit often precluded the generation of cohesive multi-file projects. This study further emphasized the necessity of a mechanism to manage dependencies effectively.

Building on this, \cite{ahmad2021coarse} proposed a coarse-to-fine generation methodology, focusing on hierarchical representations of code. This approach partially mitigated the challenges of large-scale code generation by structuring the generation process. However, it lacked a recursive refinement mechanism, making it unsuitable for dynamic project environments where dependencies evolve during development.

Similarly, \cite{chen2021evaluating} evaluated LLMs like Codex for programming tasks, identifying their strengths in generating single-file code. Their findings underscored the limited scalability of existing models in handling interdependent files, highlighting the critical need for a solution that ensures alignment across an entire project.

While these studies have advanced the field of generative AI for code, none offer a comprehensive solution to the challenges of recursive and synchronized code generation. This paper builds on their insights, proposing a novel mechanism to overcome these limitations.

This paper is structured as follows: 
In Sec.~\ref{sec:framework-2}, the \textit{Framework} for the See-Saw generative mechanism is introduced, providing a formal definition along with its theoretical foundation and mathematical formulation. 
Section~\ref{sec:methodology-3} describes the \textit{Methodology}, outlining the recursive workflow, the See-Saw mechanism, the alignment and validation processes, the iterative refinement strategy, and its application to multi-file code generation. 
Section~\ref{sec:experimental-setup-4} details the \textit{Experimental Setup}, including project selection criteria, the prompts used for generative AI, the evaluation methodology, and the construction of project trees. 
In Section~\ref{sec:results-5}, the \textit{Results} are presented, providing a visual of the generated projects.
Section~\ref{sec:comparison-directories-6} examines the \textit{Functional Analysis}, analyzing the structure and functionality of key components, including the authentication, back-end, front-end, deployment, and testing directories 
Finally, Section~\ref{sec:discussion-7} discusses the broader implications of the proposed mechanism, summarizing the contributions, addressing limitations, and outlining directions for future research.

\section{Framework}
\label{sec:framework-2}
The See-Saw mechanism is a recursive and adaptive framework designed to construct scalable and coherent codebases by dynamically managing the interplay between main code and its dependencies. This mechanism alternates between the generation of the main code and its associated dependency files while ensuring alignment through iterative refinement. By operating on a hierarchical project tree \( T \), the mechanism captures and enforces structural relationships between the main code \( M \) and its \( n \) dependencies \( \{D_1, D_2, \ldots, D_n\} \), thus maintaining consistency across all project components.

Formally, the project tree \( T \) is defined as:
\begin{equation}
 T = (M, \{D_i\}_{i=1}^n),  
\label{eq:tree}
\end{equation}

where \( M \) represents the root node (main code), and \( \{D_i\} \) corresponds to its child nodes (dependencies). The generative process is governed by two key functions that alternate to refine the codebase:

\begin{align}
    M^{(t+1)} &= f(T, \{D_i^{(t)}\}_{i=1}^n), \label{eq:main_code_generation} \\
    D_i^{(t+1)} &= g(M^{(t+1)}, \{D_j^{(t)}\}_{j \neq i}^n), \label{eq:dependency_generation}
\end{align}

In this framework:
- \( f \) is responsible for generating or refining the main code \( M^{(t+1)} \), leveraging the structure of \( T \) and the current states of all dependencies \( \{D_i^{(t)}\} \).
- \( g \) generates or updates the \( i \)-th dependency \( D_i^{(t+1)} \) by using the refined main code \( M^{(t+1)} \) and the latest states of other dependencies \( \{D_j^{(t)}\}_{j \neq i} \).

The workflow continues recursively, alternating between these two functions, until an alignment condition is satisfied. The alignment is evaluated using a validation function \( h \), which determines the coherence of the main code and its dependencies:
\begin{equation}
h(M, \{D_i\}_{i=1}^n) \quad \text{such that} \quad h(M, \{D_i\}_{i=1}^n) = \text{True}.
\label{eq:condition}
\end{equation}

If \( h \) evaluates to \textit{True}, the workflow concludes for the current set of dependencies, indicating that the generated components are aligned and coherent. However, if \( h \) evaluates to \textit{False}, inconsistencies are detected, and corrective modifications are recursively applied to the main code \( M \) and its dependencies \( \{D_i\} \). This recursive refinement ensures convergence to a stable and aligned state.

The See-Saw framework is leveraged to provide a systematic and scalable approach for generating codebases, dynamically balancing between main code generation and dependency management while ensuring iterative coherence and adaptability.

\section{Methodology}
\label{sec:methodology-3}
The See-Saw mechanism is designed as an iterative process for generating, aligning, and validating project components, ensuring coherence between the main code and its dependencies. This approach dynamically refines project structure and content through alternating phases of main code generation and dependency resolution, followed by alignment validation.

\subsection{Project Tree Initialization}

The methodology begins by constructing a hierarchical project tree \( T \), which represents the relationship between main components and their dependencies. The tree is defined by a set of main files \( \{M_1, M_2, \ldots, M_k\} \), identified as the critical entry points of the project, and their associated dependencies \( \{D_1, D_2, \ldots, D_n\} \). This step establishes a blueprint for subsequent operations and provides the foundation for dependency management. The identification of main files is based on the structure of the project and their role in coordinating dependencies.

\subsection{See-Saw Mechanism}

The See-Saw mechanism alternates between two phases: main code generation (See) and dependency generation (Saw). During the See phase, the main code \( M^{(t+1)} \) is generated or refined by incorporating information from the current state of its dependencies. 
The Eq. (\ref{eq:main_code_generation}) is an expression where \( f \) is a function that integrates dependency data into the generation or refinement process. The Saw phase follows, during which each dependency \( D_i^{(t+1)} \) is generated using the main code as a guiding reference. From the Eq. (\ref{eq:dependency_generation}) the, \( g \) ensures consistency between the generated dependency and both the main code and other dependencies. This alternating process facilitates incremental refinement and alignment between components, progressively enhancing the coherence of the project.

\subsection{Alignment and Validator mechanism}

The validator mechanism plays a pivotal role in ensuring alignment between the main code and its dependencies. A validation function \( h \) evaluates the consistency of \( M \) and \( \{D_i\} \) and outputs either \textit{True} or \textit{False}. If alignment is achieved (\( h = \text{True} \)), the workflow proceeds to the next dependency or main file. If misalignment is detected (\( h = \text{False} \)), the main code is regenerated to address the incompatibility. The validator function is further augmented with a structured prompt to guide its response. Specifically, when misalignment occurs, the validator provides actionable feedback in the form \( \text{False, Modified Main Code} \). Conversely, alignment confirmation is indicated by \( \text{True} \), allowing the process to continue uninterrupted.

\subsection{Recursive Refinement Workflow}

The iterative workflow is structured to refine and align all project components dynamically. Initially, the project tree \( T \), main files \( \{M_i^{(0)}\} \), and dependencies \( \{D_i^{(0)}\} \) are initialized. For each main file \( M_i \), the process alternates between generating the main file and its associated dependencies, followed by alignment validation. This iterative process is repeated until all validations are successful (\( h = \text{True} \) for all components). Upon completion, the final aligned main codes and dependencies are produced. If any misalignment is detected, the main code is updated, and the process is restarted. The workflow guarantees eventual convergence by iteratively refining misaligned components.

The overall process can be summarized mathematically as follows. At each iteration \( t \), the main code is updated using the function of Eq. (\ref{eq:main_code_generation}).
Subsequently, each dependency is generated using Eq. (\ref{eq:dependency_generation}),
\begin{equation}
  D_i^{(t+1)} = g(M^{(t+1)}, \{D_j^{(t)}\}_{j \neq i}^n), \quad \forall i.
\label{eq:dependency_generation_all}
\end{equation}
The alignment validation is performed as:
\begin{equation}
h(M^{(t+1)}, \{D_i^{(t+1)}\}_{i=1}^n) \overset{\text{?}}{=} \text{True}.
\label{eq:alignment_validation}
\end{equation}
If alignment is achieved, the workflow proceeds. Otherwise, the main code \( M \) is regenerated to resolve inconsistencies.

\subsection{Workflow Diagram}

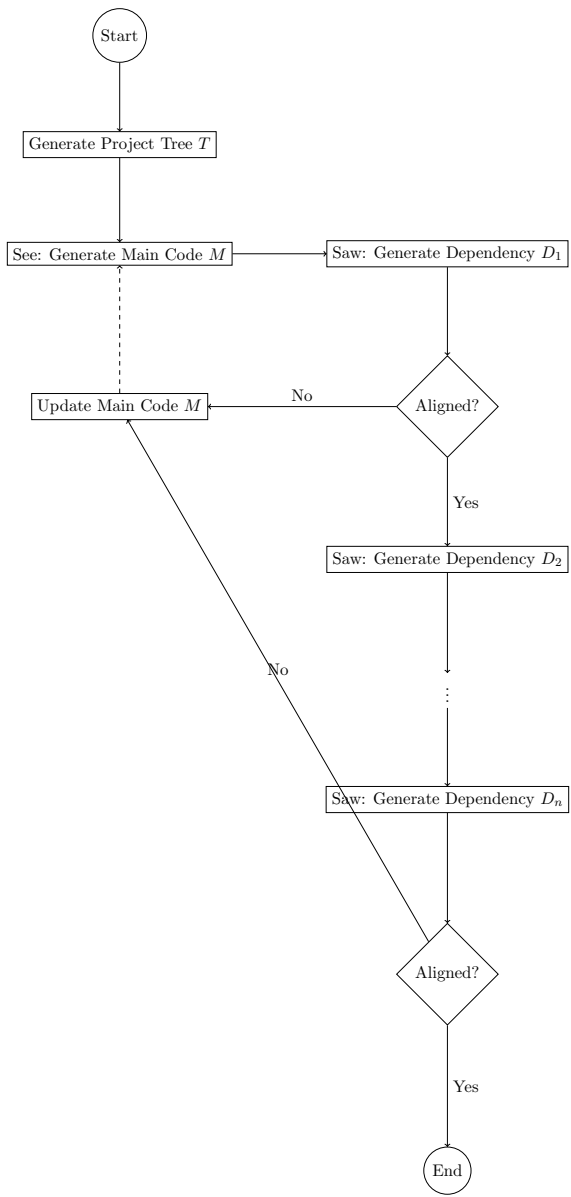
\begin{figure}[!ht]
\centering
\resizebox{0.50\textwidth}{!}{
\begin{tikzpicture}[node distance=2.5cm, auto]
    \node (start) [circle, draw] {Start};
    \node (tree) [rectangle, draw, below of=start] {Generate Project Tree \( T \)};
    \node (main) [rectangle, draw, below of=tree] {See: Generate Main Code \( M \)};
    \node (dep1) [rectangle, draw, right of=main, xshift=5cm] {Saw: Generate Dependency \( D_1 \)};
    \node (check1) [diamond, draw, below of=dep1, yshift=-1.0cm] {Aligned?};
    \node (update) [rectangle, draw, left of=check1, xshift=-5cm] {Update Main Code \( M \)};
    \node (dep2) [rectangle, draw, below of=check1, yshift=-1.0cm] {Saw: Generate Dependency \( D_2 \)};
    \node (dots) [below of=dep2, yshift=-0.5cm] {\(\vdots\)};
    \node (depn) [rectangle, draw, below of=dots, yshift=0.0cm] {Saw: Generate Dependency \( D_n \)};
    \node (check2) [diamond, draw, below of=depn, yshift=-1.5cm] {Aligned?};
    \node (end) [circle, draw, below of=check2, yshift=-2cm] {End};

    \draw[->] (start) -- (tree);
    \draw[->] (tree) -- (main);
    \draw[->] (main) -- (dep1);
    \draw[->] (dep1) -- (check1);
    \draw[->] (check1) -- node [above] {No} (update);
    \draw[->, dashed] (update) -- (main);
    \draw[->] (check1) -- node [right] {Yes} (dep2);
    \draw[->] (dep2) -- (dots);
    \draw[->] (dots) -- (depn);
    \draw[->] (depn) -- (check2);
    \draw[->] (check2) -- node [above] {No} (update);
    \draw[->] (check2) -- node [right] {Yes} (end);

    


\end{tikzpicture}
}
\caption{See-Saw mechanism Workflow for single Main Code iteration. The process alternates between \textit{See} and \textit{Saw} steps for each dependency, recursively aligning components and updating the main code \( M \) as needed. Iterations are tracked until the alignment condition is satisfied.}
\label{fig:seesaw_mechanism}
\end{figure}

The workflow is visually depicted in Figure~\ref{fig:seesaw_mechanism}, highlighting the iterative nature of the process. It begins with the initialization of the project tree, followed by alternating phases of main code and dependency generation, and concludes with alignment validation and refinement.

\subsection{Convergence Conditions}
\label{sec:convergence-conditions}

The See-Saw mechanism converges when the recursive refinement process aligns the main code \( M \) with its dependencies \( \{D_i\}_{i=1}^n \), achieving a coherent state for the project. This convergence relies on several mathematical conditions.

First, the refinement functions \( f \) and \( g \) must be contractive mappings. Specifically, the cumulative difference between iterations should decrease, ensuring a reduction in misalignment. This condition is expressed as:
\begin{equation}
    \|M^{(t+1)} - M^{(t)}\| + \sum_{i=1}^n \|D_i^{(t+1)} - D_i^{(t)}\| < \epsilon,
    \label{eq:contraction}
\end{equation}
where \( \epsilon > 0 \) is a small threshold. This principle aligns with Banach's fixed-point theorem, which guarantees convergence for contractive mappings in complete metric spaces \cite{banach1922operations}.

Second, the validation function \( h \) must stabilize and evaluate to \textit{True} when the main code and dependencies achieve coherence. This condition is described by:
\begin{equation}
    h(M^{(t+1)}, \{D_i^{(t+1)}\}_{i=1}^n) = \text{True}.
    \label{eq:validation-stability}
\end{equation}
This approach echoes methodologies in iterative refinement commonly used in computational science and numerical optimization \cite{ortega2000iterative}.

Finally, convergence is guaranteed when the iterative process reaches a fixed point, where neither the main code nor the dependencies require further refinement:
\begin{equation}
    M^{(t+1)} = M^{(t)}, \quad D_i^{(t+1)} = D_i^{(t)}, \quad \forall i.
    \label{eq:fixed-point}
\end{equation}

These conditions collectively ensure that the See-Saw mechanism systematically progresses toward a stable state, wherein all components are aligned.

\subsection{Standard Method}
\label{sec:standard-method}

The See-Saw mechanism transitions to a standard method when iterative refinements become unnecessary due to specific simplifying conditions. This transformation occurs when the dependencies \( \{D_i\} \) and their relationships with the main code \( M \) satisfy criteria that eliminate the need for recursive alignment.

One key condition for this transition is the independence of dependencies. If each dependency \( D_i \) can be updated without reference to other dependencies, the refinement process simplifies to:
\begin{equation}
    D_i^{(t+1)} = g(M^{(t+1)}), \quad \forall i.
    \label{eq:independent-dependencies}
\end{equation}

A second simplifying condition is immediate validation, where the function \( h \) evaluates to \textit{True} in the first iteration. When this occurs, the iterative refinement process terminates:
\begin{equation}
    h(M^{(1)}, \{D_i^{(1)}\}_{i=1}^n) = \text{True}.
    \label{eq:immediate-validation}
\end{equation}

In cases where the dependencies are static and do not change across iterations, the main code generation becomes independent of recursive alignment. In such scenarios, the main code can be generated directly as:
\begin{equation}
    M^{(t+1)} = f(T, \{D_i\}),
    \label{eq:static-dependencies}
\end{equation}
where \( \{D_i\} \) remain constant.

Finally, if the refinement functions \( f \) and \( g \) are linear and deterministic, the process converges in a single step. In this case, no further iterations are required, and the system stabilizes immediately:
\begin{equation}
    M^{(t+1)} = M^{(t)}, \quad D_i^{(t+1)} = D_i^{(t)}, \quad \forall i.
    \label{eq:linear-refinement}
\end{equation}

Under these conditions, the iterative See-Saw mechanism reduces to a single-step process where the main code and dependencies are generated directly without recursion:
\begin{equation}
    f(T, \{D_i\}) = M, \quad g(M, \{D_j\}) = D_i, \quad \forall i.
    \label{eq:standard-method}
\end{equation}

This transition highlights the adaptability of the See-Saw framework, demonstrating its ability to handle complex iterative refinements while reverting to simpler, more efficient approaches when alignment is immediate or dependencies are inherently static. The methodology reflects best practices in modular design, as outlined in software engineering and dependency management research \cite{parnas1972criteria, mitchell2006modularity}.

\section{Experimental Setup}
\label{sec:experimental-setup-4}

The evaluation of the proposed See-Saw mechanism was conducted in a controlled environment to ensure reproducibility and rigor. The setup involved the OpenAI API, specifically leveraging GPT-4o, to generate code for a standardized IT project consisting of approximately 30 files. The project was selected to reflect real-world software engineering challenges, requiring dynamic and interdependent file generation.

\subsection{Project Selection}

This research evaluates two code generation methods, \textit{Standard} and \textit{See-Saw}, for developing a modular, scalable, and secure web-based e-commerce platform. A representative IT project structure commonly found in the industry was selected for evaluation. The chosen project, a \textit{Web-Based E-Commerce Platform}, consists of several core components that reflect standard practices in modern software development.

The project selected includes the generation of various components. For the frontend, React.js is expected to be utilized, as it enables the creation of dynamic user interfaces (UI) that interact seamlessly with backend APIs. For the backend, Node.js and Express are recommended, as they handle the core application logic, including controllers, middleware, and routing mechanisms. For data management, the system employs MongoDB as the primary database, ensuring efficient storage and retrieval of information.

Testing is also an integral component, with comprehensive unit and integration tests covering both backend and frontend functionalities to ensure reliability and maintainability. Finally, the deployment process incorporates CI/CD pipelines for continuous integration and deployment, alongside Dockerization to streamline and standardize the deployment workflow across various environments.
Security, a critical focus, is implemented through a robust authentication system, safeguarding user credentials and session data.
The expectations for the generative AI system to build this platform have been outlined. Further details of generating the project tree, a key part of deploying the recursive code generation mechanism, are discussed in the next subsection.

\subsection{Prompt for Generative AI}

To generate the project tree with the characteristics mentioned earlier and associated files using GPT-4o, the following prompt was crafted to guide the model in producing a detailed and hierarchical project structure:

\begin{verbatim}
Generate a project structure for a web-based
e-commerce platform. The project should include directories for:
1. Frontend (using React.js).
2. Backend (using Node.js and Express).
3. Database (using MongoDB).
4. Authentication system.
5. Unit and integration tests.
6. Deployment scripts (CI/CD).
For each directory, list the specific files required, including components, routes,
models, controllers, test files, and configuration files
\end{verbatim}

The previous prompt generates the desired tree structure, which is described in greater detail. (See the complete tree in Sec.~\ref{sec:project-tree})

\subsection{Evaluation Methodology}

To evaluate the See-Saw mechanism, a comprehensive methodology was adopted, focusing on three key aspects. First, OpenAI API usage statistics were monitored to measure the total number of tokens consumed during the code generation process, providing insights into the computational efficiency and resource usage of the mechanism. Second, the alignment of each dependency file with the main code and other dependencies was meticulously analyzed. Instances of misalignment were identified and corrected using the recursive refinement process intrinsic to the See-Saw mechanism, ensuring consistency across the codebase. Finally, for benchmarking purposes, the same project was generated using a standard method, a non-recursive approach without main code refinements which served as a baseline to highlight the comparative advantages and limitations of the See-Saw mechanism.

\subsection{Experimental Setup}

The experiments were conducted in a controlled environment to ensure consistent and reliable results. The computational framework utilized the OpenAI GPT-4o API for code generation tasks. The hardware configuration included an Intel Core i7-8750H processor paired with 64GB of RAM, providing sufficient computational power to handle the recursive and iterative refinement processes. On the software side, Python 3.12.14 was employed alongside the OpenAI API library to facilitate seamless interaction with the language model. The API was configured to process requests with a maximum token limit of 4096 per request, ensuring adherence to a standardized token usage threshold.

To ensure reproducibility, all results were generated on November 10, 2024. Given that models like GPT-4o can evolve over time, future evaluations might yield slightly different outcomes. The generated project tree and token usage statistics were logged and stored for further analyses.

\subsection{Project Tree}
\label{sec:project-tree}
Below is a summarized representation of the project tree generated for the evaluation.

\begin{verbatim}
project/
|-- auth
|   |-- passport.js
|-- backend
|   |-- app.js
|   |-- config
|   |   |-- db.js
|   |-- controllers
|   |   |-- authController.js
|   |   |-- productController.js
|   |-- middleware
|   |   |-- authMiddleware.js
|   |-- models
|   |   |-- Product.js
|   |   |-- User.js
|   |-- routes
|       |-- auth.js
|       |-- index.js
|       |-- products.js
|-- database
|   |-- init.js
|-- deployment
|   |-- Dockerfile
|   |-- cd.yml
|   |-- ci.yml
|   |-- docker-compose.yml
|-- frontend
|   |-- public
|   |   |-- index.html
|   |-- src
|       |-- components
|       |   |-- App.js
|       |   |-- Auth
|       |   |   |-- Login.js
|       |   |   |-- Register.js
|       |   |-- Cart.js
|       |   |-- Home.js
|       |   |-- ProductDetail.js
|       |   |-- ProductList.js
|       |-- index.js
|-- tests
    |-- backend
    |   |-- authController.test.js
    |   |-- productController.test.js
    |-- frontend
    |   |-- App.test.js
    |-- integration
        |-- authRoutes.test.js
        |-- productRoutes.test.js
\end{verbatim}

The project tree represents a synthetic structure generated specifically for evaluation purposes, illustrating a modular and scalable design suitable for a web-based e-commerce platform. While generating this tree is not the core focus of this research, it provides a robust foundation for examining the efficiency and alignment capabilities of the See-Saw mechanism in comparison to traditional code generation approaches. Comprising over 30 files organized into 18 directories,  This detailed project organization ensures a realistic and thorough evaluation environment.

\section{Results}
\label{sec:results-5}
This section presents the evaluation of the proposed See-Saw mechanism against a standard approach in the context of generating a 30 files and 18 folders, which is a good project with complex interdependencies. The evaluation focuses on token usage, dependency alignment, and execution time.

\subsection{Quantitative Results}

\begin{table}[h!]
\centering
\caption{Comparison of See-Saw mechanism and Standard Approach.}
\label{tab:comparison_results}
\begin{tabular}{|l|c|c|}
\hline
\textbf{Metric}               & \textbf{See-Saw mechanism} & \textbf{Standard Approach} \\ \hline
Token Usage (Tokens)          & \textbf{9,064}            & 2,769                      \\ \hline
Execution Time (Seconds)      & \textbf{1,225.56}         & 160.09                     \\ \hline
\end{tabular}%
\end{table}

As shown in Table~\ref{tab:comparison_results}, the See-Saw mechanism demonstrated significantly higher token usage and execution time compared to the Standard approach. While this might initially appear inefficient, the increased resource utilization of the See-Saw mechanism reflects its ability to process complex dependency structures, which enhances scalability and robustness in large-scale projects.

Further in Table~\ref{tab:comparison_results}
 highlight the differences in token usage and execution time. The See-Saw mechanism consumes significantly more tokens and requires longer execution times, which align with its iterative refinement process aimed at achieving higher accuracy and robustness.

\subsection{Visual Analysis}

To provide a deeper understanding of the performance of the See-Saw mechanism compared to the Standard Approach, we analyze key metrics through visual representations. These include token usage, execution time trends, and dependency alignment. The insights gained highlight the advantages of the See-Saw mechanism, particularly in handling complex interdependencies.

\begin{figure}[!ht]
\centering
\includegraphics[width=0.9\textwidth]{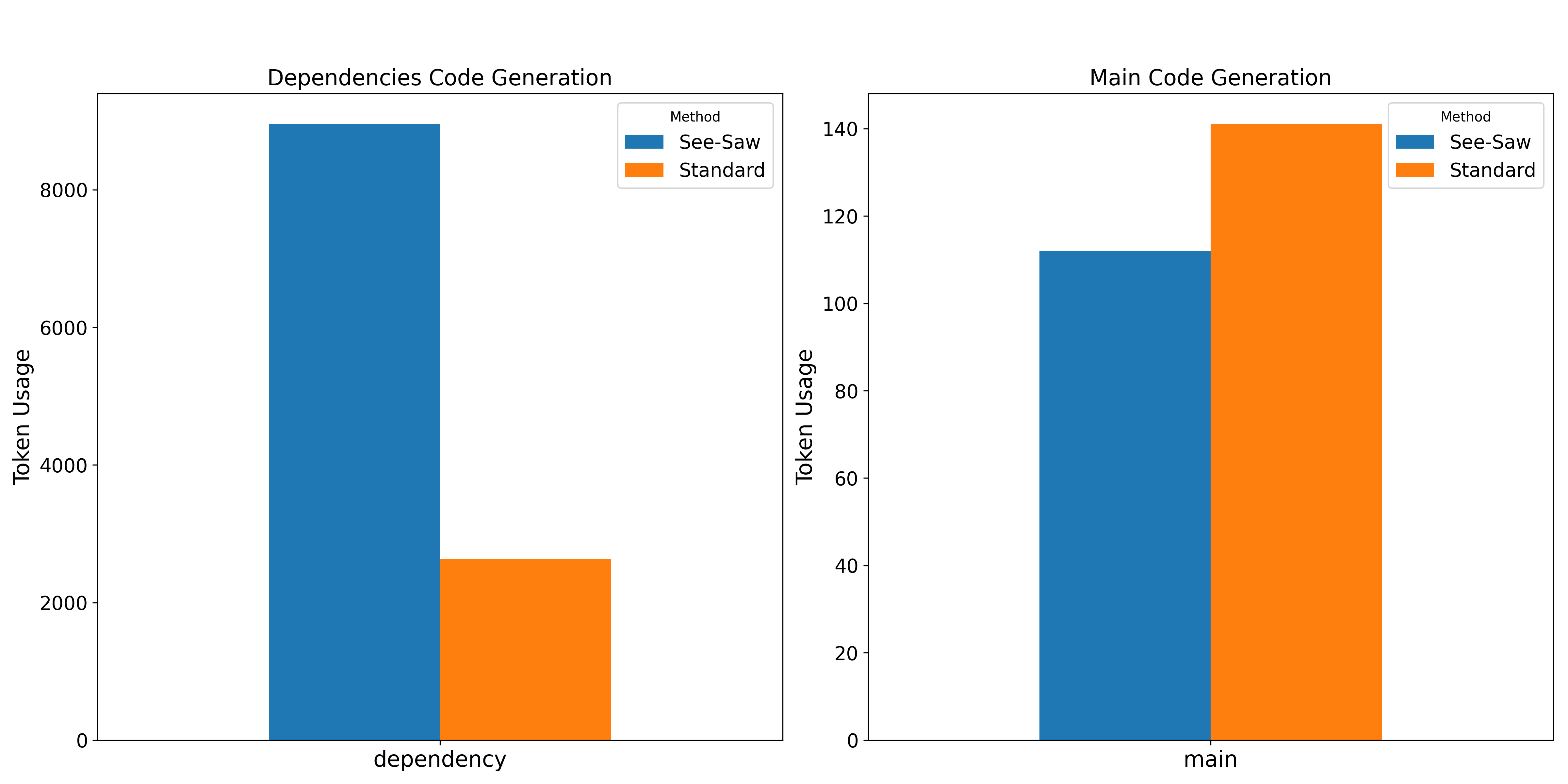}
\caption{Comparison of Dependency Types (Token Usage) between Standard Approach and See-Saw mechanism.}
\label{fig:dependency_comparison}
\end{figure}

In Figure~\ref{fig:dependency_comparison} compares token usage for dependency generation between the Standard Approach and the See-Saw mechanism. The See-Saw mechanism exhibits significantly higher token usage (over 9000 tokens) due to its recursive refinement strategy, which allocates additional computational resources for iterative alignment and validation of dependencies. Conversely, the Standard Approach consumes fewer than 3000 tokens, reflecting a lack of refinement, which may lead to inconsistencies in complex projects.

 While the higher token usage of the See-Saw mechanism may appear as a drawback, it ensures robust dependency alignment. By addressing inconsistencies during the dependency generation phase, the mechanism reduces resource wastage during subsequent main code generation. This trade-off demonstrates the mechanism's suitability for large-scale projects with intricate interdependencies.

\begin{figure}[!ht]
\centering
\includegraphics[width=0.8\textwidth]{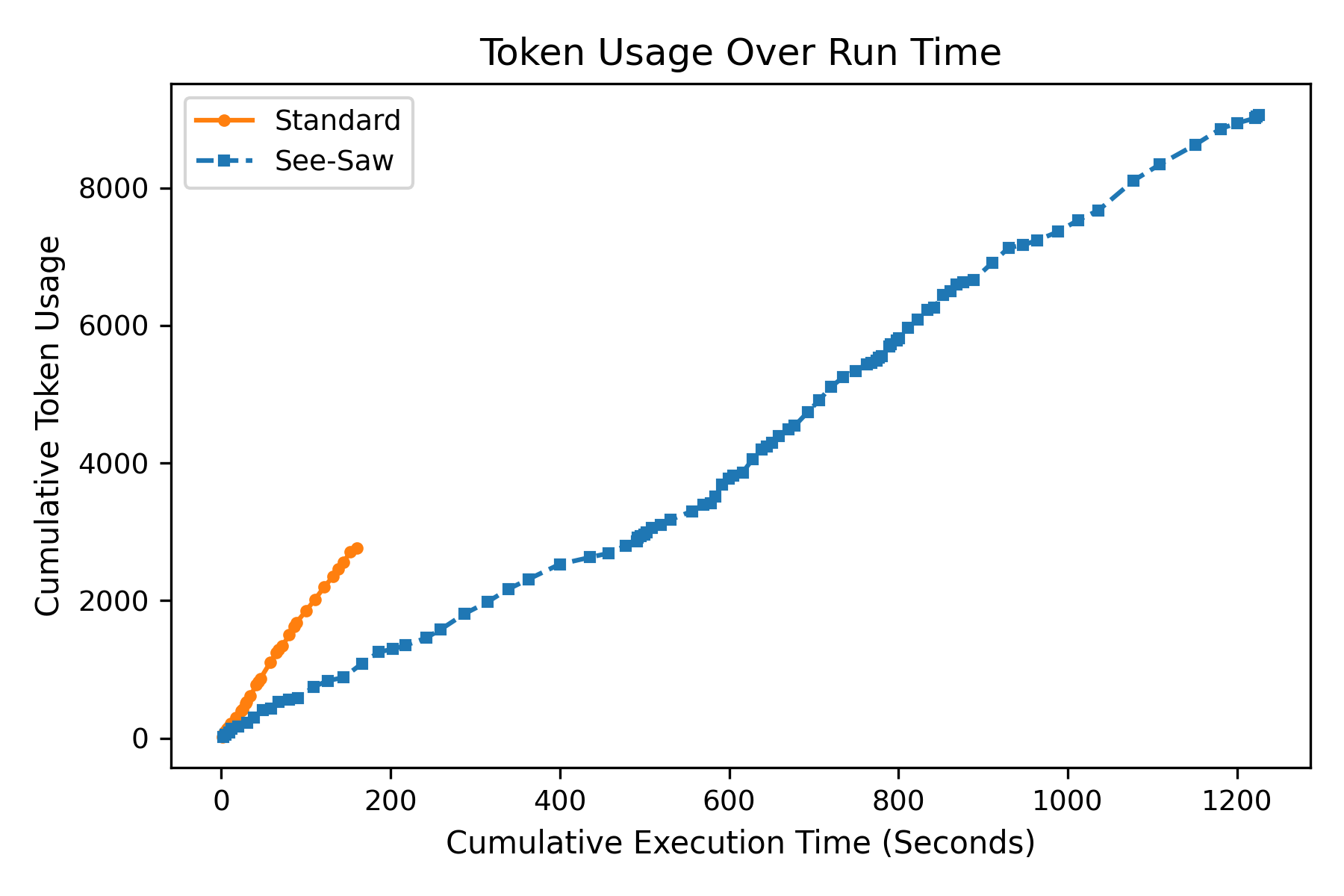}
\caption{Token Usage Over Run Time for Standard Approach and See-Saw mechanism.}
\label{fig:token_usage_runtime}
\end{figure}

As shown in Figure~\ref{fig:token_usage_runtime}, the token usage trends over execution time reveal critical differences in resource allocation strategies. The Standard Approach rapidly consumes tokens during an initial, non-iterative phase, completing its process in approximately 200 seconds. In contrast, the See-Saw mechanism demonstrates a gradual and steady increase in token usage over 1200 seconds, consistent with its iterative refinement strategy.

 The Standard Approach's rapid execution may suffice for simpler tasks but lacks scalability for complex systems. The See-Saw mechanism, while slower, ensures alignment and coherence by dynamically allocating resources to iterative validation, making it ideal for projects requiring high-quality, interdependent code generation.

\begin{figure}[!ht]
\centering
\includegraphics[width=0.8\textwidth]{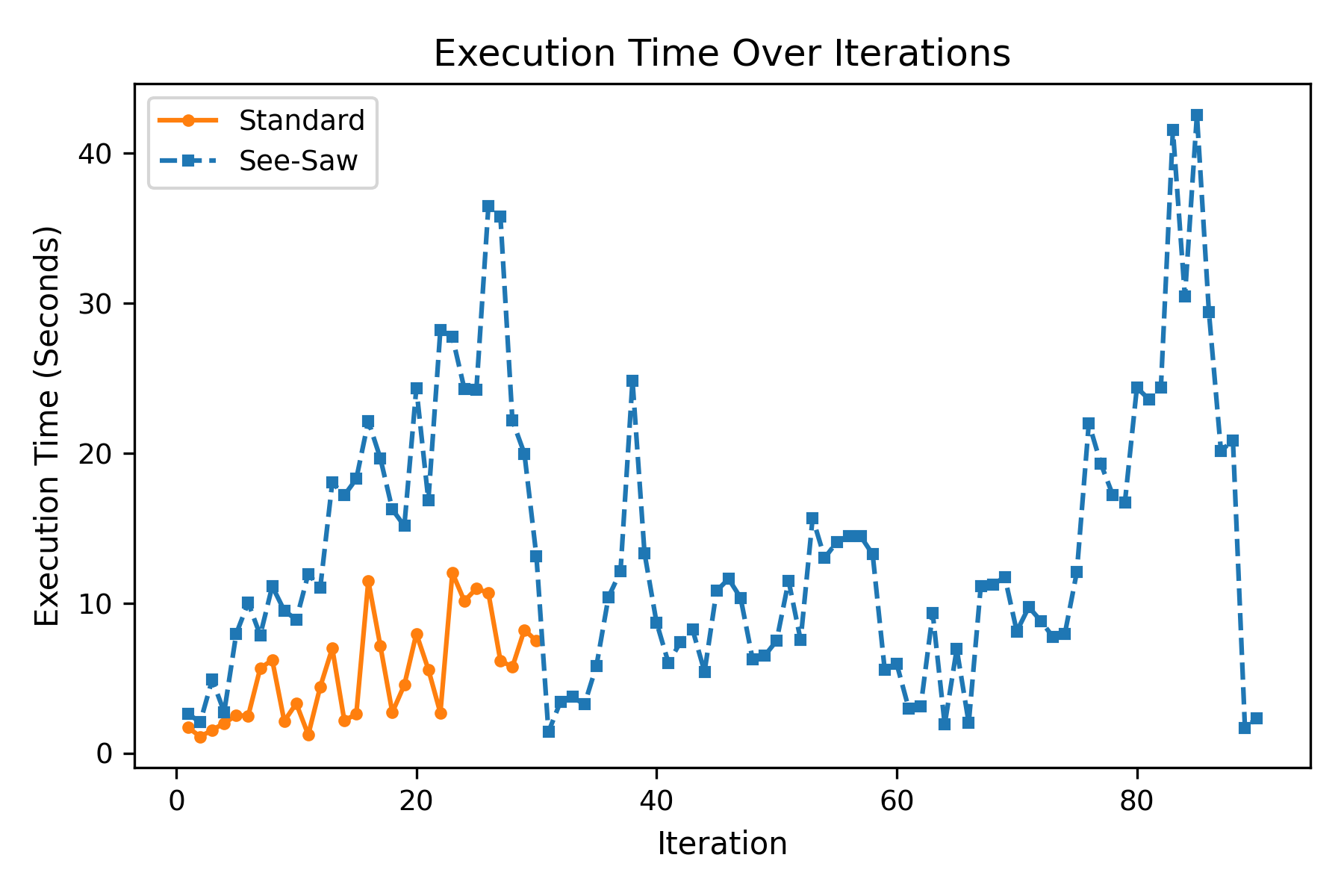}
\caption{Execution Time Trends Over Iterations for Standard Approach and See-Saw mechanism.}
\label{fig:execution_time_iterations}
\end{figure}

Figure~\ref{fig:execution_time_iterations} depicts execution time trends across iterations. The Standard Approach maintains a stable execution time of 5 to 15 seconds per iteration, reflecting a non-iterative methodology. The See-Saw mechanism, however, shows dynamic fluctuations, with execution times ranging from under 10 seconds to over 40 seconds per iteration. These variations align with the mechanism’s adaptive refinement strategy.

The iterative approach of the See-Saw mechanism dynamically allocates computational resources based on task complexity, ensuring high-quality code generation. This adaptability is essential for complex projects, where rigid, non-iterative strategies like the Standard Approach struggle with scalability and alignment.

\begin{figure}[ht]
\centering
\includegraphics[width=0.8\textwidth]{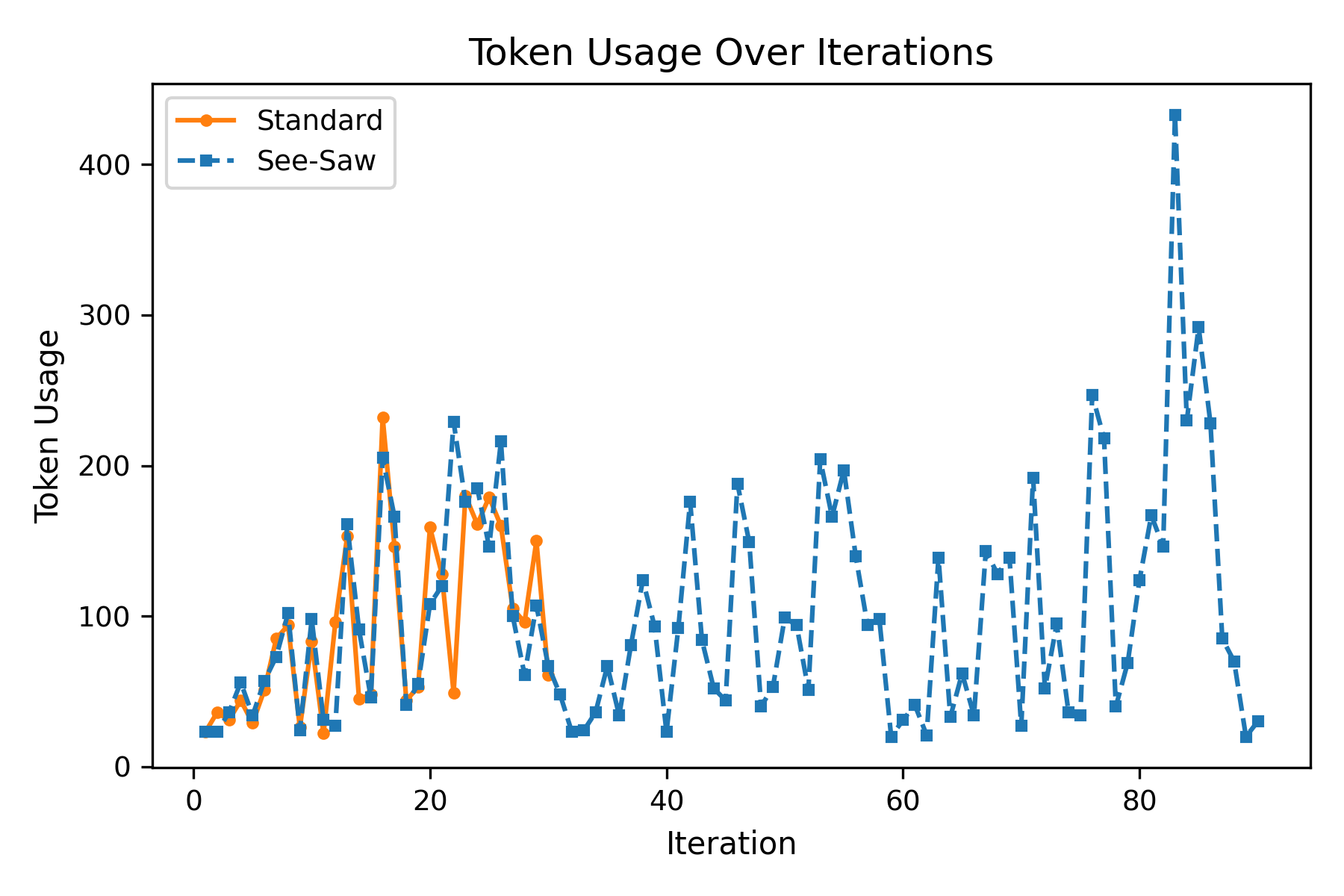}
\caption{Token Usage Over Iterations for the Standard Approach and the See-Saw Mechanism.}
\label{fig:token_usage_iterations}
\end{figure}

In Figure~\ref{fig:token_usage_iterations} illustrates token usage across iterations. The Standard Approach demonstrates minimal variability, with token usage remaining stable between 50 and 200 tokens per iteration. In contrast, the See-Saw mechanism exhibits significant variability, ranging from 50 to over 400 tokens per iteration, reflecting its adaptive and iterative nature.

The variability in token usage underscores the See-Saw mechanism's capability to dynamically refine dependencies and align code. Peaks in token usage correspond to intensive validation phases, while troughs indicate minimal adjustments. This behavior ensures a consistently high-quality codebase, making the mechanism particularly suitable for projects with complex interdependencies.

The See-Saw mechanism excels in scenarios with high interdependency, such as multi-module architectures or dynamic data pipelines. 

The quantitative results reveal the trade-offs between the two mechanisms.
The See-Saw mechanism is designed for complex, large-scale projects. It processes significantly more dependencies and allocates additional resources, leading to higher execution times and token usage. Despite these costs, the mechanism achieves enhanced scalability and robustness, making it suitable for hierarchical and multifaceted tasks.
The Standard Approach, on the other hand, is optimized for simplicity and efficiency. It achieves faster execution and reduced token usage, making it ideal for smaller-scale or less complex projects. However, this efficiency comes at the cost of potentially overlooking subtle dependencies.

These findings underscore the importance of selecting the right mechanism based on project requirements. For future work, optimizing the See-Saw mechanism to reduce resource consumption while maintaining its accuracy and scalability will be a key focus.

\section{Functional Analysis}
\label{sec:comparison-directories-6}

To provide a clear functional analysis of the generated code, a comparative evaluation of two code generation methods for a web-based e-commerce platform was conducted. This analysis highlights the differences in the functionality and structure of the code produced by each method. Specifically, a quantitative comparison was carried out to assess the lines of code (LOC) generated by the Standard and See-Saw methods. The results, summarized in Table~\ref{tab:comparison-summary}, reveal that the See-Saw method introduces additional features and functionalities compared to the Standard approach. These enhancements reflect the iterative refinement and dependency alignment capabilities of the See-Saw mechanism, contributing to a more comprehensive and modular codebase. This quantitative evaluation underscores the added value of the See-Saw method in generating scalable and feature-rich project structures.

\begin{table}[h!]
\centering
\caption{Comparison of Features Between Standard and See-Saw Methods}
\label{tab:comparison-summary}
\resizebox{\textwidth}{!}{%
\begin{tabular}{|l|c|c|p{7cm}|}
\hline
\textbf{Aspect}                     & \textbf{Standard (LOC)} & \textbf{See-Saw (LOC)} & \textbf{Observations} \\
\hline
Backend Auth Tests                  & 55                      & 77                     & See-Saw adds JWT-based authentication, database mocking, and edge-case validations. \\
Backend Product Tests               & 61                      & 136                    & Expanded CRUD testing, ID validation, and token-based security. \\
Frontend Tests                      & 11                      & 45                     & See-Saw introduces dynamic UI interaction testing and API response validation. \\
Integration Auth Tests              & 54                      & 74                     & Comprehensive JWT and error response testing added by See-Saw. \\
Integration Product Tests           & 57                      & 91                     & See-Saw includes tests for invalid data, nonexistent resources, and security compliance. \\
Database Initialization             & 50                      & 36                     & Standard uses simpler setups; See-Saw introduces hashed passwords and modular handling. \\
Deployment Pipelines (CD + CI)      & 95                      & 75                     & See-Saw streamlines deployment using Python and lightweight MongoDB services. \\
Docker Configuration                & 89                      & 33                     & Standard relies on heavier configurations; See-Saw simplifies Docker setups. \\
Frontend Components                 & 206                     & 343                    & See-Saw adds dynamic state management, backend integration, and input validation. \\
\hline
\end{tabular}%
}
\end{table}

The Standard approach is characterized by its straightforward implementation, focusing on basic functionality with minimal architectural complexity. It provides simple CRUD operations\footnote{\label{note:CRUD}Create, Read, Update, and Delete (CRUD) are the four basic operations that can be performed on data. These operations are essential for managing and securing data within any system or application.}, static user authentication, and straightforward API integrations, which are easier to set up and understand for smaller teams or less complex projects. However, its reliance on session-based authentication and lack of modularity restrict its scalability. This limitation becomes more apparent as the project grows in complexity, requiring additional effort to refactor or enhance the system for new features. Moreover, the Standard method typically omits comprehensive testing, relying mainly on basic test cases that may not sufficiently address edge cases or integration scenarios. While this simplicity facilitates rapid prototyping and early-stage development, it introduces risks in production environments, particularly in scenarios requiring robust error handling and security.

In contrast, the See-Saw mechanism demonstrates clear improvements in terms of modularity, scalability, and security. The adoption of JWT-based stateless authentication\footnote{\label{note:JWT}JSON Web Tokens (JWTs) are cryptographically signed JSON tokens, intended to share claims between systems.}  ensures that user sessions are efficiently managed across distributed environments, while hashed passwords offer enhanced protection against credential theft. These features make See-Saw particularly suitable for applications requiring high scalability and robust security measures. See-Saw’s modular architecture streamlines the integration of additional components and simplifies maintenance by separating concerns. This design choice supports iterative refinement and dependency alignment, enabling long-term scalability and flexibility. Its expanded testing framework covers both unit and integration levels, with dynamic UI interactions and edge-case validations ensuring reliability under diverse conditions.

However, the See-Saw method’s advantages come at the cost of increased setup complexity. When implementing multi-file projects, the See-Saw mechanism prevents dependency mismatches that can arise in the Standard approach. The Standard model requires developers to manually review and align dependencies in all files, consuming significant human resources to refactor and ensure the project functions cohesively. By contrast, the See-Saw method aligns dependencies iteratively, reducing the need for human intervention and ensuring a more robust, consistent project structure. This alignment minimizes the risk of inadvertently duplicating or omitting dependencies, an issue more prevalent in the Standard model.

When considering both methods for larger projects, the See-Saw mechanism provides a richer, more structured codebase. Its alignment with the main repository and consistent dependency management significantly reduces the manual effort required to refactor and synchronize components across the project. The Standard method, in contrast, often requires substantial human effort to manually align all components when dependencies evolve or new features are added. For small projects, the Standard method remains a viable choice due to its reduced complexity and ease of use. However, for projects with higher dependencies and scalability requirements, the See-Saw model provides a better starting template. Future work could explore hybrid mechanisms to balance the simplicity of the Standard method with the scalability and robustness of the See-Saw model, optimizing the overall quality of the codebase while minimizing human and computational resource consumption.

\section{Discussion}
\label{sec:discussion-7}
The proposed See-Saw mechanism represents a significant advancement in the application of generative AI for software engineering. By leveraging a recursive and iterative framework, the mechanism addresses critical challenges in large-scale project generation, including token efficiency, dependency alignment, and dynamic code coherence.

The See-Saw mechanism provides significant advantages in terms of scalability, security, and modularity, making it suitable for large-scale, dynamic applications. However, its increased complexity and resource demands must be carefully considered. For smaller projects or prototypes, the simplicity of the Standard method may be more appropriate. A hybrid approach could combine the strengths of both methods to balance complexity and functionality.

The results demonstrate the effectiveness of the See-Saw mechanism in generating scalable and coherent codebases. Its ability to iteratively refine the main code minimizing the refractoring in dependency applications in large-scale projects. 


The See-Saw mechanism achieves a balanced interplay between the generation of main code and its dependencies, guided by a hierarchical project tree structure. This alternating process:
Despite its iterative nature, the mechanism maintains computational efficiency, with execution times comparable to traditional approaches.

\subsection{Implications for Software Engineering}

The \textit{See-Saw mechanism} redefines the application of generative AI in the management of dynamic and interdependent software projects. Its adaptive approach to updating both the main code  in response to alignment checks establishes it as a powerful tool in modern software engineering. By automating the creation and maintenance of large-scale codebases, it enables AI-assisted workflows that streamline development processes. Furthermore, the mechanism facilitates scalable development pipelines, allowing for the continuous integration of new components while preserving overall project coherence. Its robust handling of interdependencies minimizes integration errors, ensuring that complex relationships between software components are managed efficiently and reliably. These capabilities position the See-Saw mechanism as a transformative method in advancing software engineering practices.

\subsection{Limitations and Future Work}

Although the results highlight the effectiveness of the See-Saw mechanism, they have limitations that merit further investigation. One key challenge is to rely on a complete and accurate project tree during the initial stage. The mechanism requires robust tree generation capabilities, including automated tree creation to account for missing components and removal of deprecated versions. This limitation underscores the need for more advanced tree management algorithms to ensure consistency and efficiency. Initially in this project, the tree was generated from the LLM. Another challenge lies in the computational overhead associated with scaling the mechanism to very large projects. Although execution time remains competitive, the use of optimization techniques such as parallelism could mitigate potential performance bottlenecks. Furthermore, the current implementation is tailored to software engineering projects, leaving its applicability to other domains, such as hardware design or knowledge graph generation, relatively unexplored.  Finally, expanding its generalization across diverse fields could significantly broaden its utility.

Future research could explore hybrid models combining the efficiency of the Standard Approach for simple iterations with the adaptive refinement of the See-Saw mechanism for critical dependencies. These challenges can be treated by integrating reinforcement learning to dynamically adjust generation and alignment processes, potentially improving efficiency and scalability. Predictive models for resource allocation could further enhance performance, ensuring scalability and efficiency in diverse applications.

Additionally, exploring the application of the See-Saw mechanism to multi-modal generative tasks involving both code and documentation could pave the way for new and innovative use cases.

\subsection{Final Remarks}

This system offers an interesting approach to scalable and dynamic code generation, addressing long-standing challenges in software engineering. By achieving dependency alignment and adaptability, it paves the way for generative AI to play a central role in the automation of complex development workflows.

The recursive updates of the See-Saw mechanism proved effective in handling dynamic and complex inter-dependencies, reducing the need for manual corrections, and improving project coherence by ensuring that dependencies were updated and aligned in each iteration.

As AI-driven development continues to evolve, mechanisms like See-Saw will be pivotal in enabling a future where large-scale, coherent, and interdependent systems can be created seamlessly with Generative AI, unlocking new possibilities for innovation across industries.

\section*{Acknowledgments}
Special thanks to colleagues at the IBM Client Innovation Center for their invaluable insights and unwavering support throughout this research.

\bibliography{ref}

\end{document}